\pdfoutput=1
\documentclass[11pt]{article}

\usepackage{EACL2023}
\usepackage{booktabs}
\usepackage{times}
\usepackage{latexsym}
\usepackage{float}
\usepackage[export]{adjustbox}
\usepackage{multirow}
\usepackage{xcolor}

\usepackage[T1]{fontenc}
\usepackage[utf8]{inputenc}

\usepackage{microtype}

\usepackage{inconsolata}

\title{Pairing Orthographically Variant Literary Words to Standard Equivalents Using Neural Edit Distance Models}

\author{Craig Messner \\
  Center for Digital Humanities\\
  Johns Hopkins University\\
  \texttt{cmessne4@jhu.edu} \\\And
  Tom Lippincott \\
  Center for Digital Humanities\\
  Johns Hopkins University\\  
  \texttt{tom.lippincott@jhu.edu} \\}

\begin{document}
\maketitle
\begin{abstract}
We present a novel corpus consisting of orthographically variant words found in works of 19\textsuperscript{th} century U.S. literature annotated with their corresponding "standard" word pair. We train a set of neural edit distance models to pair these variants with their standard forms, and compare the performance of these models to the performance of a set of neural edit distance models trained on a corpus of orthographic errors made by L2 English learners. Finally, we analyze the relative performance of these models in the light of different negative training sample generation strategies, and offer concluding remarks on the unique challenge literary orthographic variation poses to string pairing methodologies.
\end{abstract}

\section{Introduction}

Using orthographic information to pair similar strings from a list of variants has a number of uses, from spelling correction to cognate detection. Beyond character level similarity, what it means to be a "good" neighbor to a given source string might entail phonetic similarity (in the case of many misspellings), some sort of cognitive proximity (in certain "point" misspellings, say "k" for "c") or it may reflect a shared linguistic history (in the case of cognates). Using orthographic information to achieve a string-pair ranking that incorporates these axes of meaning requires that the orthography of a string also captures this other meaningful dimension, to a greater or lesser degree. We introduce the domain of "literary orthographic variants," a set of orthographic modifications motivated by literary aesthetic concerns instead of purely linguistic or cognitive principle, and posit that the orthographic results of these unique motivations necessitates re-evaluating modeling approaches that have proven successful in more linguistically motivated domains (such as the aforementioned cognate detection and spelling correction tasks). We evaluate this claim using string pairing, a task where a model compares two strings and outputs a probability that they are a match. These strings can be considered a match if their pair probability exceeds a certain threshold. Furthermore, these probabilities can also be used to rank a set of possible string pairs. We obtain these probabilities using a neural edit distance model architecture, an approach that has proven effective at pairing cognate words. Specifically we train neural edit distances models on a corpus of nonliterary orthographic variants produced by L2 English learners and a novel corpus of literary variants in order to offer the following contributions:

\begin{itemize}
    \item The aforementioned novel corpus of literary orthographic variants that can support training and evaluation of string-pairing models
    \item  Analysis of this corpus and how its specific set of challenges differ from datasets of orthographic variants that are not derived from literary sources
    \item  Evaluation of the impact of negative example generation strategies on model performance across different domains
    \item Initial steps towards a general system able to pair literary orthographic variants to their standard forms
    
\end{itemize}

\section{Background}

\subsection{Literary Orthographic Variation}
While orthographic variation is often framed by deviance from an accepted standard, it has also been used in a literary context as a vehicle of meaning. This technique is notably prevalent in the literature of the 19\textsuperscript{th} century United States, where it often served to identify a particular character as belonging to a certain race, class, gender or region \cite{ives1971theory} \cite{jones1999strange}.
Buoyed by English orthography's highly redundant nature \cite{shannon1951prediction} the presence of topic-specific surrounding context, and the desire to have the variation itself be meaningful (perhaps by using a particular system of orthography that signifies a certain subject position) literary orthographic variants are typically more extreme and more obscurely motivated than those produced as the result of misspellings or other similar processes. 

\subsection{String Pairing Using Edit Distance Methods}
Edit distance measures, most commonly the Levenshtein Distance \cite{levenshtein1966binary} have been used to rank variant-standard token pairs. More recently, statistical edit distance \cite{ristad1998learning} and neural edit distance \cite{libovicky-fraser-2022-neural} have allowed edit distance to be learned empirically from data. While statistical edit distance learns a single distribution of edit operations over paired strings, neural edit distance uses a differentiable version of the expectation maximization (EM) algorithm as a loss function for a neural model.  This allows neural edit distance to learn edit operation probabilities from contextual embeddings. \cite{libovicky-fraser-2022-neural} train a neural edit distance string pairing model that employs RNN learned embeddings and randomly generated negative samples in order to achieve state of the art performance on a cognate detection task \cite{rama-etal-2018-automatic}.

\section{Methods and Materials}

\subsection{Project Gutenberg Corpus\footnote{The full corpus is available \href{https://github.com/comp-int-hum/edit-distance-ml}{here}}}
We first use the Project Gutenberg (GB) catalog file\footnote{available \href{https://www.gutenberg.org/ebooks/offline_catalogs.html}{here}} to subset the full collection to English texts produced by authors living in the 19\textsuperscript{th} century. We then limit this set to those works identified as part of the Library of Congress "PS" (American Literature) classification group. We tokenize each of this subset of texts and split into sentences before automatically identifying possible orthovariant tokens using a variety of criteria, including:
\begin{itemize}
    \item Presence of numeric characters
    \item Presence of capitalization
    \item Presence of candidate token in WordNet \cite{miller1995wordnet} or the Brown Corpus \cite{francis1964standard}
\end{itemize}
We sampled sentences with possible orthovariant tokens randomly, and Author 1 provided standard token annotations for the tokens deemed actually variant. The final corpus consists of 3058 variant tokens paired with their standard variants and their sentence-level context. 

\subsection{FCE Corpus}
The Cambridge Learner First Certificate in English (FCE) corpus is comprised of short narratives produced by English as a second language (ESL) learners \cite{yannakoudakis-etal-2011-new}. The corpus includes hand tagged corrections for a variety of observed linguistic errors. We subsetted the corpus to only include errors with a possible orthographic component, indicated by the "S" class of error codes \cite{nicholls2003cambridge}. This resulted in a subset of 4757 samples.

\subsection{Empirical Characterization of Corpora}
\begin{table}[h!]
\centering
\tabcolsep=0.13cm
\begin{tabular}{lllll}
\hline
\textbf{Corpus} & \textbf{1LD\%}  & \textbf{2LD\%} & \textbf{3LD\%} &
\textbf{4+LD\%} \\
\hline
FCE  & 74.1  & 20.9 & 3.2 & 1.8 \\
     
Gutenberg & 43.8  & 28.9 & 17.2 & 10.1 \\

% HT   & 52.4 & 30.7 & 11.0 & 5.9 \\
\hline
\end{tabular}
\caption{\label{ap1}Levenshtein distances of standard and nonstandard tokens in tagged samples, expressed as percentage.}
\end{table}
Consistent with our hypothesis about the differences between literary and nonliterary orthovariants, Table~\ref{ap1} demonstrates that the nonstandard tokens found in GB tend to be more distant from their "standard" pairings. This empirically demonstrates at least one axis of difference between the GB corpus and corpora commonly used to evaluate approaches to string pairing, alignment and ranking. 

\begin{table*}[t]
\centering
\begin{tabular}{llrrrr}
\toprule
 &  & \textbf{F-FCE} & \textbf{F-GB} & \textbf{MRR-FCE} & \textbf{MRR-GB} \\
\textbf{Model} & \textbf{Count} &  &  &  &  \\
\midrule
\multirow[t]{4}{*}{LD} & 10 & 0.81 & 0.69 & 0.40 & 0.29 \\
 & 20 & 0.79 & 0.66 & 0.64 & 0.34 \\
 & 30 & 0.76 & 0.60 & 0.67 & 0.41 \\
 & 50 & 0.72 & 0.56 & 0.63 & 0.44 \\
\cline{1-6}
\multirow[t]{4}{*}{mixed} & 10 & 0.84 & 0.72 & 0.59 & 0.52 \\
 & 20 & 0.81 & 0.67 & 0.65 & 0.57 \\
 & 30 & 0.79 & 0.68 & 0.68 & 0.56 \\
 & 50 & 0.77 & 0.62 & 0.67 & \color{red}0.62 \\
\cline{1-6}
\multirow[t]{4}{*}{random} & 10 & 0.97 & 0.93 & 0.61 & 0.47 \\
 & 20 & 0.97 & 0.93 & 0.65 & 0.53 \\
 & 30 & 0.96 & 0.90 & 0.69 & 0.52 \\
 & 50 & 0.94 & 0.87 & \color{blue}0.70 & 0.50 \\
\bottomrule
\end{tabular}

\caption{The blue highlighted cell is the best performing model trained on the FCE corpus, red is the best performing trained on the GB corpus. F scores indicate each model's ability to distinguish true and false string pairs, MRR scores indicates the ability of each model to rank a set of standard-variant pairs generated using Brown}
\label{table:Perf}
\end{table*}

\subsection{Experiment 1: Neural Edit Distance String Pairing for Candidate Filtering\footnote{Code for the following experiments is available at \url{https://github.com/comp-int-hum/edit-distance-ml}}}
We train a neural edit distance model on a string pairing task and empirically derive a probability threshold in order to separate likely variant/standard token pairs from unlikely pairs. We generate negative samples by pairing variant observed tokens with tokens drawn from Brown using the following methods:
\begin{enumerate}
    \item Random: \textit{n} randomly selected known false tokens sourced from Brown
    \item LD: \textit{n} lowest LD from source variant known false tokens
    \item Mixed: \textit{n}/2 Random process tokens and \textit{n}/2 LD process tokens
\end{enumerate}
We perform this procedure for \textit{n} of 10, 20, 30 and 50. We split the data into test, train and validation sets, each containing a (necessarily unique) admixture of known positive and known negative generated pairs. Following the method of \cite{libovicky-fraser-2022-neural} we generate a match probability threshold for each model during training by adjusting it to maximize evaluation F1 score, and then evaluated each model's ability to distinguish true and false token pairings also using F1 score.

\subsection{Experiment 2: Neural Edit Distance String Pairing for Pair Prediction}
Leaving aside negative generate pairs, we pair each known true source token in a given test set with all of the tokens found in Brown. We then employ the models trained in Experiment 1 to rank the probability of each source-Brown pairing being a true pair. We evaluate the accuracy of these rankings using mean reciprocal rank (MRR). 

\section{Results and Analysis}
Results of the experiments can be found in Table \ref{table:Perf}. The F-score of each experiment necessarily depends on the unique set of negatives generated by a given count (10, 20 etc.) and technique (LD, random, mixed). As it might be expected, models given the more difficult task, in whole or part, of distinguishing low LD variants perform worse. However, the inclusion of these difficult pairs seem to benefit GB's performance when it comes to overall pair ranking. These MRR scores (columns MRR-FCE and MRR-GB) are produced using only Brown and a stable test set of known positive pairs in each given corpus, and thus form the basis of our comparison.

\begin{figure}[H]
\includegraphics[width=.45\textwidth]{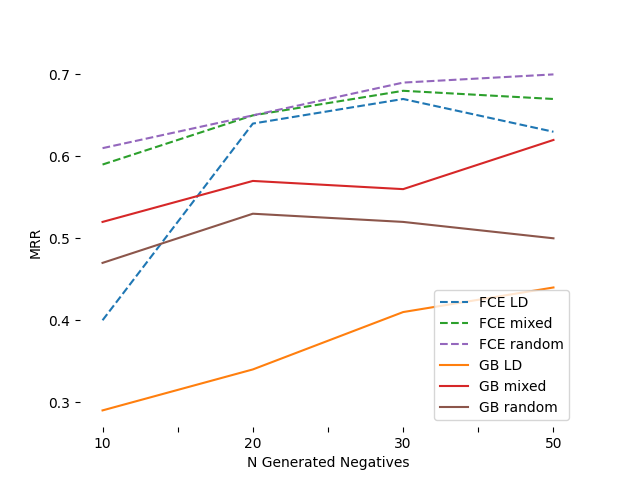}
\caption{MRR by N Generated Negatives}
\label{fig:MRR}
\end{figure}

Figure \ref{fig:MRR} shows that for all \textit{n} of negative samples, the models trained on FCE perform best when provided negative samples generated by the random process. The models employing close LD negatives performed uniformly the worst. This is somewhat the opposite of our expected result. The negative signal of incorrect close LD examples would on its face seem particularly useful given the near LD nature of FCE's spelling and usage errors, as distinguishing between the candidates in the near LD neighborhood of a variant token becomes imperative.

On the other hand, models trained on GB perform uniformly the best when provided negatives generated by the mixed strategy, a combination of random and close LD pairs. This implies that in the case of GB, but not FCE, the two sources of negative examples provide orthogonal information that are each of their own particular use during the training process. 

\begin{figure}[H]
\includegraphics[width=.45\textwidth]{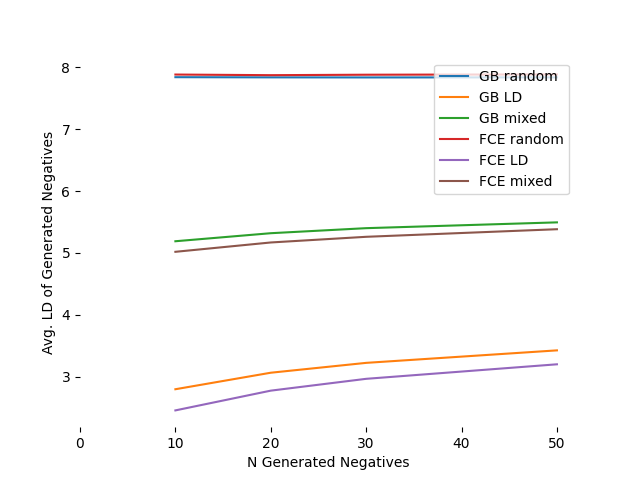}
\caption{Average LD of Generated Negatives}
\label{fig:LDNeg}
\end{figure}

The specific character of the generated negative examples may explain this performance disparity. Figure \ref{fig:LDNeg} shows the average LD from the target variant tokens for each the of negative generation strategies. Random generation, the best performing strategy over FCE, produces negative samples on average about 8 LD from the target variant token, no matter the number generated. Logically, the mixed strategy, which performed best over GB, produced a set of samples with average an LD falling between the uniformly high LD of the randomly generated samples and the low LD of the samples generated by the LD process, which, for GB range from just below to just over 3 LD on average. 
\begin{table*}[t]
\begin{tabular}{l|llllll}
\hline
\textbf{Standard} & \textbf{Variants} &            &              &          \\
\hline
afraid  & afear'd   &   avraid  &   'feerd  &   'fraid  & 'afeared &    ofraid  \\
children          & childens          & child'n    & chillunses   & chilther \\
master            & mars’             & mars’r     & 'marse       & mauster  \\
convenient        & convanient        & conwenient & conuenenient &          \\
office            & awffice           & oflfis     & ohfice       &  
\\
calculate         & calkylate         & calkelate  & ca'culate    &          \\
\hline
\end{tabular}
\caption{Samples of paired standard and variant wordforms found in GB}
\label{table:Samples}
\end{table*}

In short, FCE trained models benefit most from uniformly high LD negative examples, while GB trained models benefit most from a mixture of high and low LD negative examples. This may speak to the distinct nature of the positive examples found in these corpora. The FCE corpus is comprised of samples produced by multiple authors. However, the range of possible orthovariant forms they employ is limited by their shared intent to adhere to a standard form of English orthography as best as they can. This overriding principle could lead FCE's variant forms to conform more closely to a centralized set of possible edits, typified by common character substitutions or phonetic misspellings -- it would be understandable for a writer making a good faith attempt at producing standard English orthography to replace a "c" with a "k", but never, say, an elision apostrophe ("'"). If this is the case, much of the information the model would need to distinguish between low LD Brown candidate tokens is already contained in the fairly uniform set of possibilities demonstrated by the positive examples -- the types of transformations embodied by these examples closely resembles the set of transformations resident in the FCE corpus as a whole.

In contrast, the GB corpus contains samples drawn from multiple authors who each employ their own looser set of orthographic constraints. These authors do not attempt in good faith to adhere to a particular standard orthography. Rather, they use orthography as an expressive tool, and may not rely as heavily on further orthographic principles. Consequently, the positive examples may lose some significant amount of explanatory value. Examples of this effect drawn from the corpus can be found in Table \ref{table:Samples}.

Each variant is a phonetic or pseudo-phonetic rendering of a given word in a form of particularly motivated variant English orthography, yet each set of character-level substitutions varies to a large degree.\footnote{It should be noted and acknowledged that many of these examples are due to the proliferation of explicitly racist depictions of African-Americans and other minority groups in this period of literature} Indeed, even though all of these forms are relatively low LD from their standard token, the set of transformation principles encoded in one teaches us relatively little about the set found in any of the others. This could explain the the mixed strategy's superior performance on GB, as the positive examples under-determine the space of likely transformations among low LD candidates, leaving the generated low LD negative examples more room to provide useful information.

\section{Future Work}
The complexities inherent in literary variant orthography offers many axes on which to continue these studies. Further experiments could be performed to validate the hypothesis concerning the mixed strategy's success on the GB corpus. This could be accomplished a number of ways, including training an additional set of models on a dataset of orthovariants generated by other means (for example the Z\'eroe character level adversarial benchmark \cite{eger2020hero}) and evaluating the performance of the negative generation strategies in the context of that dataset's own orthographic precepts. 

Additionally, future work could leverage the sentence-level contextual information included in the GB corpus to aid in string pair ranking. This could be an especially fruitful solution given the multiply-systematic nature of literary orthographic variation, as local information about the semantics of the source variant and the nature of the orthographic choices made in nearby neighbor tokens could aid in adjudicating between source-candidate pairs granted relatively similar probabilities by the neural edit distance model. 
\newpage
%\section*{Acknowledgements}
\bibliography{custom.bib, anthology}
\bibliographystyle{acl_natbib}

\appendix
\label{sec:appendix}

\section{Hyper-parameters and model details}
\label{sec:hypers}
The hyperparameters we employ closely follow those found in \cite{libovicky-fraser-2022-neural}. The RNN embedding model employs gated recurrent units (GRU) \cite{cho2014learning}. The model was trained using three equally weighted loss functions:  EM, binary cross entropy, and non-matching negative log-likelihood.

\begin{tabular}{lr}
\multicolumn{2}{c}{ } \\
\hline
Name & Value \\
\hline
Embedding model & RNN\\
Embedding size & 256 \\
Hidden layers & 2 \\
Batch size & 512 \\
Validation frequency & 50\\
Patience & 10 \\
\hline
\end{tabular}
\vspace{1cm}

\end{document}